\documentclass[11pt,a4paper]{article}
\usepackage[hyperref]{eacl2021}
\usepackage{times}
\usepackage{latexsym}

\usepackage{amsmath}
\usepackage{enumitem}
\usepackage{hyperref}
\usepackage{graphicx}
\usepackage{wasysym}
\usepackage{multicol}
\usepackage{multirow}
\usepackage{booktabs}
\usepackage{float}
\usepackage{color}
\usepackage{diagbox}
\usepackage{xcolor,colortbl}
\newcommand{\bc}{\cellcolor{lightgray}}

\usepackage{microtype}

\aclfinalcopy 
\def\aclpaperid{284} 

\title{FakeFlow: Fake News Detection by Modeling the \\Flow of Affective Information}

\author{
Bilal Ghanem$^{1}$, Simone Paolo Ponzetto$^{2}$, Paolo Rosso$^{1}$, Francisco Rangel$^{3}$ \\
  {\normalsize $^{1}$Universitat Politècnica de València, Spain}\\
  {\normalsize $^{2}$University of Mannheim, Germany} \\
  {\normalsize $^{3}$Symanto Research, Germany} \\
  \small{ bigha@doctor.upv.es, simone@informatik.uni-mannheim.de,}\\ 
  \small{ prosso@dsic.upv.es, francisco.rangel@symanto.com}
}

\date{}

\begin{document}
\maketitle
\begin{abstract}
  Fake news articles often stir the readers' attention by means of emotional appeals that arouse their feelings. Unlike in short news texts, authors of longer articles can exploit such affective factors to manipulate readers by adding exaggerations or fabricating events, in order to affect the readers' emotions.
  To capture this, we propose in this paper to model the flow of affective information in fake news articles using a neural architecture. The proposed model, FakeFlow, learns this flow by combining topic and affective information extracted from text. We evaluate the model's performance with several experiments on four real-world datasets. The results show that FakeFlow achieves superior results when compared against state-of-the-art methods, thus confirming the importance of capturing the flow of the affective information in news articles.
\end{abstract}

\section{Introduction}
\label{sec:intro}
In today's information landscape, fake news are used to manipulate public opinion \cite{zhou2018fake} by reshaping readers' opinions regarding some issues. In order to achieve this goal, authors of fake news' narratives need to capture the interest of the reader. Thus, they are putting efforts to make their news articles look more objective and realistic. This is usually done by adding misleading terms or events that can have a negative or positive impact on the readers’ emotions.

Short text false information, e.g., fake claims or misleading headlines, might be less harmful than news articles. They may have some eye-catching terms that aim to manipulate the readers' emotions \cite{chakraborty2016stop}. In many cases, the identification of this kind of exaggeration in short statements can unmask the fabrication. On the other hand, in fake news articles the authors exploit the length of the news to conceal their fabricated story. This fact exposes the readers to be emotionally manipulated while reading longer texts that have several imprecise or fabricated plots. The flow of information has been investigated for different tasks: \newcite{reagan2016emotional} studied the emotional arcs in stories in order to understand complex emotional trajectories; \newcite{maharjan2018letting} model the flow of emotions over a book and quantify its usefulness for predicting success in books; \newcite{kar2018folksonomication} explore the problem of creating tags for movies from plot synopses using emotions.

Unlike previous works \cite{rashkin2017truth,shu2018fakenewsnet,castelo2019topic,ghanem2020emotional} that discarded the chronological order of events in news articles, in this work we propose a model that takes into account the affective changes in texts to detect fake news. We hypothesize that fake news has a different distribution of affective information across the text compared to real news, e.g. more fear emotion in the first part of the article or more overall offensive terms, etc. Therefore, modeling the flow of such information may help discriminating fake from real news. Our model consists of two main sub-modules, topic-based and affective information detection. We combine these two sub-modules since a news article's topic may have a correlation with its affective information. For example, a fake news article about Islam or Black people is likely to provoke fear and express negative sentiment while another fake news that is in favor of a particular politician might try to evoke more positive emotions and also express some exaggerations. \\
\noindent
The contributions of our work are as follows:

\begin{itemize}[leftmargin=4mm]
    \item We design a model that detects fake news articles by taking into account the flow of affective information\footnote{Available at \href{https://github.com/bilalghanem/fake_flow}{https://github.com/bilalghanem/fake\_flow}}.
    \item Extensive experiments on four standard datasets demonstrate the effectiveness of our model over state-of-the-art alternatives.
    \item We build a novel fake news dataset, called MultiSourceFake, that is collected from a large set of websites and annotated on the basis of the joint agreement of a set of news sources.
\end{itemize} 

\section{Related Work}

Previous work on fake news detection is mainly divided into two main lines, namely with a focus on social media \cite{zubiaga2015towards,aker2017simple,ghanem2019upv} or online news articles \cite{tausczik2010psychological,horne2017just,rashkin2017truth,barron2019proppy}. In this work we focus on the latter one. Fact-checking \cite{karadzhov2017fully, zlatkova2019fact, shu2019defend} is another closely related research topic. However, fact-checking targets only short texts (that is, claims) and focuses on using external resources (e.g. Web, knowledge sources) to verify the factuality of the news.
The focus in previous work on fake news detection is mainly on proposing new feature sets. \newcite{horne2017just} present a set of content-based features, including readability (number of unique words, SMOG readability measure, etc.), stylistic (frequency of part-of-speech tags, number of stop words, etc.) and psycholinguistic features (i.e., several categories from the LIWC dictionary \cite{tausczik2010psychological}). When these features are fed into a Support Vector Machine (SVM) classifier and applied, for instance, to the task of distinguishing satire from real news, they obtain high accuracies.
Using the same features for the task of fake news detection, however, results in somewhat lower scores.  \newcite{perez2018automatic} propose a model (FakeNewsDetector) that uses a feature set consisting of unigrams and bigrams, psycholinguistic, readability, punctuation and dependency-based syntactic features, and they evaluate the performance of their model in a cross-domain experiment. \newcite{rashkin2017truth} use a model based on ngram features with a Max-Entropy classifier and apply it to a dataset with different types of fake news articles (e.g., satire, hoax, propaganda, etc.). Similar to the previous work, the authors evaluate their system's performance on in-domain and out-of-domain test sets, respectively.
News, and in particular fake news, are dynamic in nature and change constantly. In order to approach the dynamic nature of news, \newcite{castelo2019topic} propose a topic-agnostic model (TopicAgnostic) that is based on morphological (count of part-of-speech tags), psycholinguistic (personal concerns, affection, and perception categories from the LIWC dictionary), readability (Gunning Fog metric, etc.) and Web-Markup features to capture patterns of the Web pages' layout (frequency of advertisements, presence of an author name, etc.). All of the morphological, psycholinguistic and readability features in the TopicAgnostic model were extracted from headlines and texts of the news articles. The approach obtains a better performance than FakeNewsDetector on three different datasets using a SVM classifier. FakeNewsTracker \cite{shu2019fakenewstracker} is a deep neural network-based model that consists of two branches: one encodes news article texts and the other encodes social media engagements (e.g., tweets and their replies). A similar model called Emotionally Infused Network (EIN) is proposed in \newcite{ghanem2020emotional}. 
EIN encodes the text of the article and their affective content, based on several dictionaries, and then combines the two vector representations. 
The authors evaluate their model on a multi-class false information dataset and show the effectiveness of using emotion features extracted from the text. 
Despite the large variety of features and models that have been explored in previous work, none of these works considers the sequence of affective information in text; instead, they feed the entire news articles as one segment into their models. In contrast, the aim of our work is to evaluate this source of information, using a neural architecture.

\section{The FakeFlow Model}

Given an input document, the FakeFlow model first divides it into $N$ segments. Then it uses both word embeddings and other affective features such as \textit{emotions, hyperbolic words}, etc. in a way to catch the flow of emotions in the document. The model learns to pay attention to the flow of affective information throughout the document, in order to detect whether it is fake or real. 

\begin{figure}
\centering
\includegraphics[width=7.6cm]{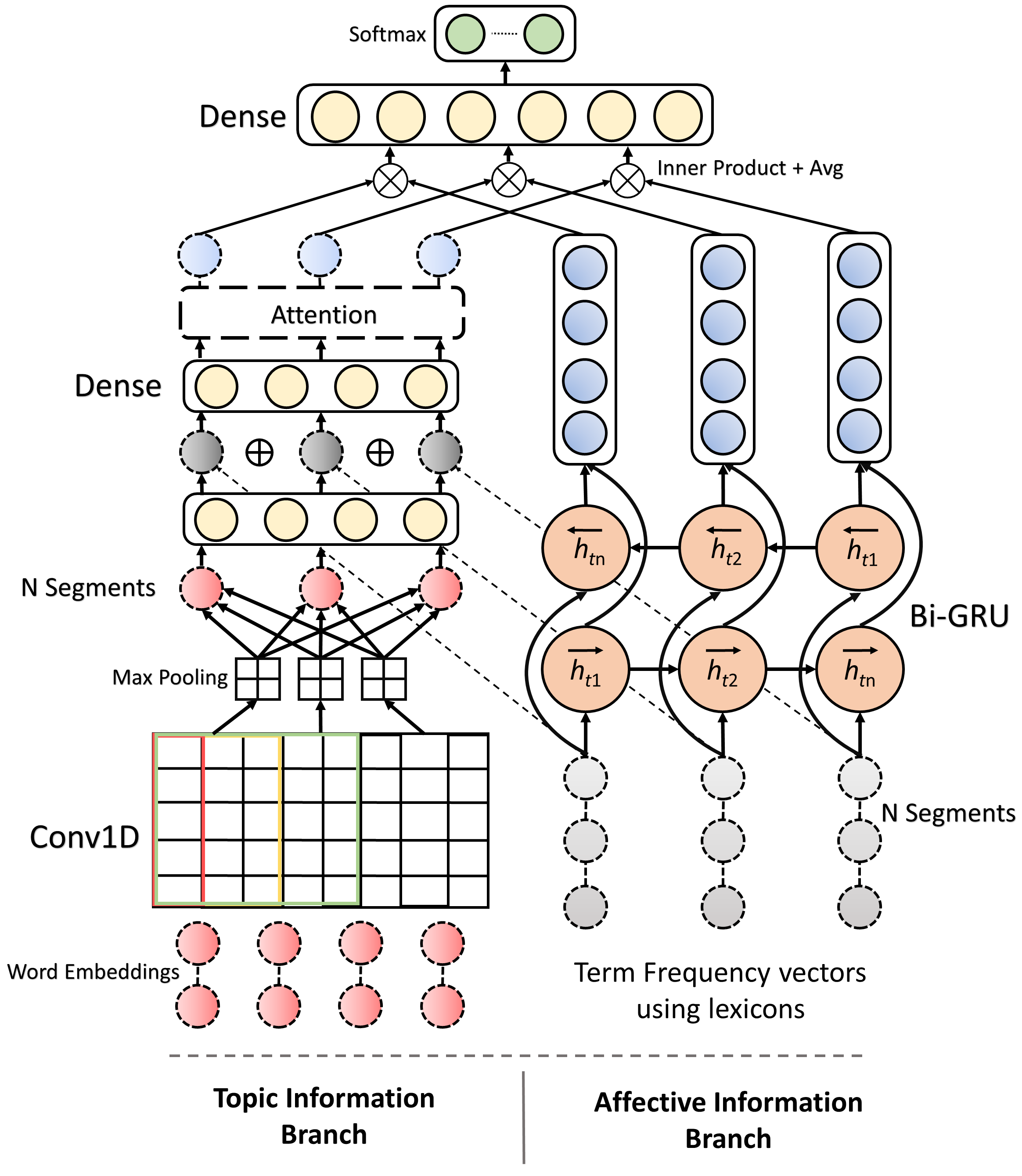}
\caption{The architecture of the FakeFlow model.}
\label{fig:fakeness_flow}
\end{figure}

Figure \ref{fig:fakeness_flow} shows the architecture of the FakeFlow model. The neural architecture has two main modules: The first module uses a Convolutional Neural Network (CNN) to extract topic-based information from articles (left branch). The second module models the flow of the affective information within the articles via Bidirectional Gated Recurrent Units (Bi-GRUs) (right branch).

\subsection{Topic-based Information}

Given a segment $n \in N$ of words, the model first embeds words to vectors through an embedding matrix. Then it uses a CNN that applies convolution processes and max pooling to get an abstractive representation of the input segment. This representation highlights important words, in which the topic information of the segment is summarized. Then it applies a fully connected layer on the output segments to get a smaller representation ($v_{\mathit{topic}}$) for later concatenation with the representation of affective information:
%
\begin{equation*}
v_{\mathit{topic}} = f(W_a \: cnn_v + b_a)
\end{equation*}
where $W_a$ and $b_a$ are the corresponding weight matrix and bias terms, and $f$ is an activation function such as ReLU, tanh, etc.

Key to FakeFlow is its ability to capture the relevance of the affective information with respect to the topics. For this, we concatenate the topic summarized vector $v_{\mathit{topic}}$ with the representation vector $v_{\mathit{affect}}$, aimed at capturing the affective information  extracted from each segment (Section \ref{sbsection:flow_info}).
%
\begin{equation*}
v_{concat} = v_{\mathit{topic}} \oplus v_{\mathit{affect}}
\end{equation*}
%
%
To merge the different representations and capture their joint interaction in each segment, the model processes the produced concatenated vector $v_{concat}$ with another fully connected layer:
%
\begin{equation*}
v_{fc} = f(W_c \: v_{concat} + b_c)
\end{equation*}
In order to create an attention-focused representation of the segments to highlight important ones and to provide the model with the ability to weight segments differently according to the similarity of neighboring segments, the model applies a context-aware self-attention mechanism \cite{zheng2018opentag} on $v_{fc}$. This is a crucial step, as the importance of a segment at timestep $t$ is related to the other segments since they share the same context in the news article. Moreover, applying the attention layer can help us understand which features are most relevant by showing to which words the network attends to during learning. The output of the attention layer is an attention matrix $l_{t}$ with scores for each token at each timestep.

\subsection{Affective Flow of Information}
\label{sbsection:flow_info}
To model the affective information flow in the news articles, we choose the following lexical features, under the assumption that they have a different distribution across the articles' segments. We use a term frequency representation weighted by the articles' length to extract the following features from each segment $n$:

\begin{itemize}[leftmargin=4mm]
    \item \textit{Emotions}: We use emotions as features to detect their change among articles' segments. For that we use the NRC emotions lexicon \cite{mohammad2010emotions} that contains $\sim$14K words labeled using the eight Plutchik's emotions (\textit{8 Features}).
    
    \item \textit{Sentiment}: We extract the sentiment from the text, \textit{positive} and \textit{negative}, again using the NRC lexicon \cite{mohammad2010emotions} (\textit{2 Features}).
    
    \item \textit{Morality}: We consider cue words from the Moral Foundations Dictionary\footnote{\href{https://moralfoundations.org/other-materials/}{https://moralfoundations.org/other-materials/}} \cite{graham2009liberals} where words are assigned to one (or more) of the following categories: \textit{care, harm, fairness, unfairness (cheating), loyalty, betrayal, authority, subversion, sanctity} and \textit{degradation} (\textit{10 Features}).
    
    \item \textit{Imageability}: We use a list of words rated by their degree of abstractness and imageability\footnote{\href{https://github.com/ytsvetko/metaphor/tree/master/resources/imageability}{https://github.com/ytsvetko/metaphor/tree/master/\\resources/imageability}}. These words have been extracted from the MRC psycholinguistic database \cite{wilson1988mrc} and then using a supervised learning algorithm, the words have been annotated by the degrees of abstractness and imageability. The list contains 4,295 and 1,156 words rated by their degree of abstractness and imageability, respectively (\textit{2 Features}).
    
    \item \textit{Hyperbolic}: We use a list of $\sim$350 hyperbolic words \cite{chakraborty2016stop}, i.e., words with high positive or negative sentiment (e.g., terrifying, breathtakingly, soul-stirring, etc.). The authors extracted these eye-catching words from clickbaits news headlines (\textit{1 Feature}).
\end{itemize}

To model the flow of the above features, we represent each segment of an article by a vector $v_{\mathit{affect}}$ capturing all 23 features listed above. Then we feed the document's vectors to a Bi-GRU network to summarize the contextual flow of the features from both directions\footnote{During prototyping, GRU produced better overall results than LSTM.} to obtain $v_{\mathit{flow}}$.

Given the segments' flow representation ($v_{\mathit{flow}}$) of an article and their relevance to the topics ($l_t$), FakeFlow applies a dot product operation and then averages the output matrix across the segments to get a compact representation $v_{\mathit{compact}}$,
%
%
which is then fed into a fully connected layer:

\begin{equation*}
v_{\mathit{final}} = f(W_d \: v_{\mathit{compact}} + b_d)
\end{equation*}
Finally, to generate the overall factuality label of an article, a softmax layer is applied to the output of the fully connected layer.

\section{Fake News Datasets}
\label{sec:datasets}

Despite the recent efforts for debunking online fake news, there is a dearth of publicly available datasets. Most of the available datasets are small in size (e.g., the Politifact\footnote{\href{https://www.politifact.com/}{https://www.politifact.com/}} dataset in \cite{shu2018fakenewsnet} has $\sim$700 available articles, the Celebrity dataset in \cite{perez2018automatic} has $\sim$500 articles, etc.), their test parts have not been manually annotated, or have been collected from a very small number of news sources. Nonetheless, we evaluate FakeFlow on three different available datasets to demonstrate its performance. In addition, we create our own dataset. Table \ref{datasets} gives an overview of the datasets that we used in our work.

\vspace{1em} \noindent \textbf{MultiSourceFake}: We rely on different resources for creating the training and test portions of the dataset, so as to provide a challenging benchmark. 

For the training part, we use \textit{OpenSources.co} (OS), \textit{MediaBiasFactCheck.com} (MBFC), and PolitiFact\footnote{\href{https://www.politifact.com/article/2017/apr/20/politifacts-guide-fake-news-websites-and-what-they/}{https://www.politifact.com/article/2017/apr/20/politifacts-guide-fake-news-websites-and-what-they/}} news websites' lists. OS list contains 560 domains, MBFC list has 548 domains, and the PolitiFact list has 227 domains.
These lists have been annotated by professional journalists. The lists contain domains of online news websites annotated based on the content type (as in the OS news list: \textit{satire, reliable}, etc.;  and in the PolitiFact news list: \textit{imposter, parody, fake news}, etc.) or from a factuality perspective (as in the MBFC news list: low, medium, and high factuality). From the OS list, we select domains that are in one of the following categories: \textit{fake, bias, reliable, hate, satire,} or \textit{conspiracy}. We consider domains under the \textit{reliable} category as real news sources, and the rest as fake. The PolitiFact list is different from the OS list since it has only labels for domains that are either fake or with mixed content. We discard the mixed ones\footnote{The discarded label is ``Some fake stories''.} and map the remaining ones to the fake news label. Finally, we select from the MBFC list those domains that are annotated either as high or low factual news and we map them to real and fake labels, respectively. Out of these three final lists, we select only those domains for our dataset that are annotated in all lists in a consistent way; for example, we discard those domains that are annotated as real in the OS list but their label in the MBFC list is fake (low factuality). The final list contains 85 news websites. We now proceed by projecting the domain-level ground truth onto the content of those domains and randomly sample articles, with a maximum of 100 news articles per domain.\footnote{Some of the websites included less than 100 news articles.}

\begin{figure}
\centering
\includegraphics[width=8cm]{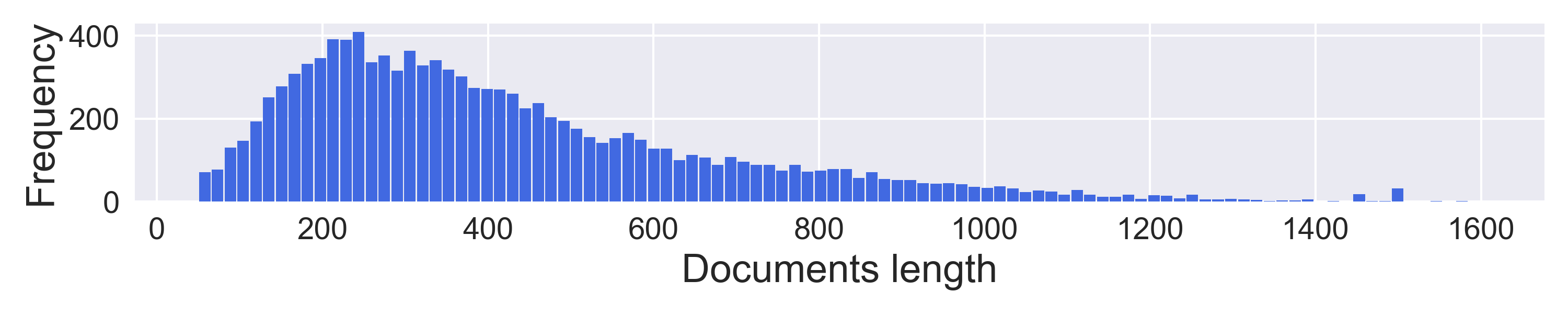}
\caption{The distribution of the documents' length in the MultiSourceFake dataset.}
\label{fig:data_len}
\end{figure}

For the test part, we use the \textit{leadstories.com} fact checking website for which professional journalists annotated online news articles on the article level as fake or real. We do not follow the way we annotate the training part since the projection of the domain-level ground truth inevitably introduces noise. The journalists that annotated \textit{leadstories.com} assigned a set of labels to the fake news articles like, e.g., \textit{false, no evidence, satire, misleading}, etc.; we map them all to the \textit{fake} label. In addition, we discard all articles that are multimedia-based. After collecting the news articles, we postprocess them by discarding very short articles (less than 30 words). The test part includes 689 fake news articles. We complement the set with a sample of 1,000 real news articles from the training part. The overall dataset consists of 5,994 real and 5,403 fake news articles. The average document length (number of words) in the MultiSourceFake dataset is 422 words, and the 95th percentile value is 942. Figure \ref{fig:data_len} shows the distribution of the documents' length in the dataset.

\vspace{1em} \noindent \textbf{TruthShades}: This dataset has been proposed in \newcite{rashkin2017truth}. The dataset was crawled from a set of domains that are annotated by professional journalists as either {\em propaganda, hoax, satire}, or {\em real}.  The dataset has been built from the English Gigaword corpus for real news, and other seven unreliable domains that annotated in one of the three previous false information labels.

\vspace{1em} \noindent \textbf{PoliticalNews}: Due to the fact that: ``a classifier trained using content from articles published at a given time is likely to become ineffective in the future'' \cite{castelo2019topic}, the authors of this work collected a dataset by crawling news websites in between the years 2013 to 2018 in order to evaluate their model's performance on different years.

\vspace{1em} \noindent \textbf{FakeNewsNet}: is a fake news repository that consists of two comprehensive datasets, one collected using claims from PolitiFact and the other from the GossipCop fact checking website. Given the large number of true and false claims from these two fact checking websites,  \newcite{shu2018fakenewsnet} built news datasets that contain visual and textual news articles content and social media information by searching Twitter for users who shared news. Out of the whole collected information, we use only the textual information of news articles, which is the part we are interested in.

\begin{table}
    \small
    \begin{tabular}{|c|c|c|c|}
    \hline
    \textbf{Name}  & \textbf{Total} & \textbf{Training} & \textbf{Test} \\ \hline
    MultiSourceFake & 11,397 & 9,708 & 1,689 \\ \hline
    TruthShades     & 23,000 & 16,000 & 4,000 - 3,000 \\ \hline
    PoliticalNews   & 14,240 & 11,392 & 2,848 \\ \hline
    FakeNewsNet     & 20,208 & 16,156 & 4,039 \\ \hline
    \end{tabular}
    \caption{Number of articles in the datasets.}
    \label{datasets} 
\end{table}

\section{Experiments}
\label{sec:experiments}

\noindent \textbf{Experimental setup.} 
We split the articles' text into $N$ segments and set the maximum length of segments to 800 words, applying zero padding to the ones shorter than 800 words. Concerning the FakeFlow hyper-parameters, we tune various parameters (\textit{dropout, the size of the dense layers, activation functions, CNN filter sizes and their numbers, pooling size, size of the GRU layer, and the optimization function}) (see Appendix \ref{sec:appendix} for the search space) using early stopping on the validation set. In addition to these hyper-parameters, we also use the validation set to pick the best number of segments ($N$). Regarding the MultiSourceFake dataset, we use 20\% of the training part for validation. 
We represent words using pre-trained word2vec \textit{Google-News-300}  embeddings\footnote{\href{https://code.google.com/archive/p/word2vec/}{https://code.google.com/archive/p/word2vec/}}. 
For evaluation, we follow the setup from related work. We report accuracy and weighted precision, recall and F1 score, and macro F1 for some datasets where the classes are imbalanced.

\vspace{1em} \noindent \textbf{Baselines.} To evaluate the performance of our model, we use a combination of fake news detection models and deep neural network architectures:

\begin{itemize}[leftmargin=4mm]
    \item \noindent \textbf{CNN, LSTM}: We use CNN and LSTM models and validate their performance when treating each document as one fragment. We experiment with different hyper-parameters and report results for the ones that performed best on the validation set.

    \item \textbf{HAN}: The authors of \cite{yang2016hierarchical} proposed a Hierarchical Attention Networks (HAN) model for long document classification. The proposed model consists of two levels of attention mechanisms, i.e., word and sentence attention. The model splits each document into sentences and learns sentence representations from words. 

    \item \textbf{BERT}: is a text representation model that showed superior performance on multiple natural language processing (NLP) benchmarks \cite{devlin2019bert}. We use the pre-trained \textit{bert-base-uncased} version which has 12-layers and yields output embeddings with a dimension of size 768. We feed the hidden representation of the special [CLS] token, that BERT uses to summarize the full input sentence, to a softmax layer. Experimentally, we found that fine-tuning BERT layers gives a higher performance. It is worth mentioning that BERT input length is limited to 512 word pieces (sub-words level) \cite{devlin2019bert}, thus, we discard the rest of the text in long news articles.
    
    \item \textbf{Fake News Detection Models}: We compare our model to several fake news detection models. We use \newcite{horne2017just} model, FakeNewsDetector \cite{perez2018automatic}, \newcite{rashkin2017truth} model, and EIN \cite{ghanem2020emotional}.\footnote{We only compare TopicAgnostic on the dataset the authors proposed (PoliticalNews).}
    
    \item \textbf{Longformer}: Giving that Transformer-based models (i.e. BERT) are unable to process long sequences, we use Longformer \citep{beltagy2020longformer}, which is a SOTA model for long document tasks. In our experiments, we set the max sequence length to 1500 to handle documents that have more than 512 tokens in the MultiSourceFake dataset (see Figure \ref{fig:data_len}). Also, we found that fine-tuning the Longformer model gives better results and a much faster convergence.
\end{itemize}
 
\section{Results and Analysis}
\label{sec:results}

\begin{figure}
\includegraphics[width=8cm]{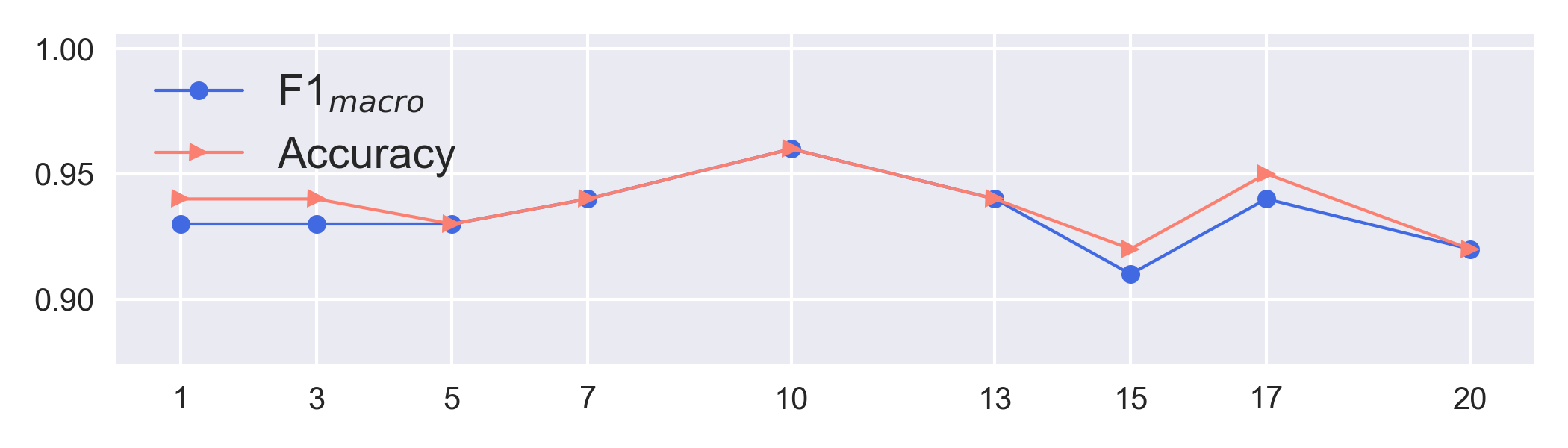}
\caption{The accuracy and F1 results of the FakeFlow model using different $N$ (number of segments).}
\label{fig:chunk_size}
\end{figure}

Table \ref{tab:results} presents the results of our proposed model and the baselines on the MultiSourceFake dataset. Our best result was achieved by using 10 as the number of segments ($N$, as found on the validation data). In Figure \ref{fig:chunk_size} we show the model's performance for segments of different length.\footnote{In the case of N=1 in Figure \ref{fig:chunk_size}, we set the maximum segment length to 1500 words instead of 800 to not lose parts of the longer articles.} In general, the results show that models that are based on either word ngrams or word embeddings are performing better than other models that use handcrafted features, e.g. \citet{horne2017just}. Also, despite the huge amount of data used to train the BERT model, the results show that BERT performs worse than \textit{FakeFlow} and also fails to outperform some of the other models. We speculate that this is due to the fact that the input length in BERT is limited to 512 words, as we mentioned previously, and a large portion of the news articles in the MultiSourceFake dataset has a length greater than 512 words. The results of the Longformer model confirm our claim regarding the documents' length and show a significantly higher F1 score than the BERT model. This emphasizes that despite the strong performance of BERT on multiple NLP benchmarks, it is unable to handle long text documents, in contrast, e.g., to vanilla text categorization \cite{adhikari2019docbert}. In addition, Longformer's results show a higher F1 score than the FakeFlow model, yet, the difference is statically insignificant.

To isolate the contribution of topical vs.\ affective information we run two simplified versions of our architecture, each consisting of the networks to capture topical and affective information only. The results show that the flow of the affect information has a weak performance when used alone; this emphasizes that affective information of a news article is a meaningful, yet complementary source of information.

\begin{table}
    \small
    \centering
    \begin{tabular}{|c|c|c|c|c|}
    \hline
    \textbf{Model} & \textbf{Acc.} & \textbf{Prec.} & \textbf{Rec.} & \textbf{F1$_{macro}$} \\
    \hline
    Majority Class                   & 0.59 & 0.35 & 0.59 & 0.37 \\ \hline
    {\scriptsize \citet{horne2017just}} & 0.80 & 0.75 & 0.78 & 0.80 \\ \hline
    FakeNewsDetector                 & 0.86 & 0.86 & 0.86 & 0.86 \\ \hline
    LSTM                             & 0.91 & 0.86 & 0.91 & 0.90 \\ \hline
    CNN                              & 0.91 & 0.89 & 0.89 & 0.91 \\ \hline
    {\scriptsize \citet{rashkin2017truth}}& 0.92 & 0.92 & 0.92 & 0.92 \\ \hline
    BERT                             & 0.93 & 0.93 & 0.94 & 0.93‡ \\ \hline
    EIN                              & 0.93 & 0.94 & 0.93 & 0.93‡ \\ \hline
    HAN                              & 0.94 & 0.94 & 0.94 & 0.93‡ \\ \hline
    Longformer                       & 0.97 & 0.97 & 0.97 & \textbf{0.97}† \\ \hline 
    
    \hline
    FakeFlow                         & 0.96 & 0.93 & 0.97 & 0.96 \\ 
    \hline 
    \scriptsize{FakeFlow -- Topic only}          & 0.91 & 0.89 & 0.90 & 0.90 \\ \hline
    \scriptsize{FakeFlow -- Affective only}         & 0.61 & 0.38 & 0.60 & 0.40 \\
    \hline
    \end{tabular}
    \caption{Results on the MultiSourceFake dataset. (‡) indicates a statistically significant improvement of \textit{FakeFlow} over the referred model using McNemar test; (†) indicates no statistically significant improvement over \textit{FakeFlow}.}
    \label{tab:results} 
\end{table}

\vspace{1em}
\noindent
\textbf{Performance on Multiple Datasets.} In Table \ref{tab:datasets} we compare the performance of the FakeFlow model to SOTA results on the other datasets we introduced in Section \ref{sec:datasets}. The TruthShades dataset has two test sets, in-domain and out-of-domain. In the in-domain configuration, training and test articles come from the same sources, and from different sources in out-of-domain configuration. The results demonstrate that FakeFlow achieves a better F1 on both test sets. In a similar way, the results on the PoliticalNews dataset show that FakeFlow also outperforms the TopicAgnostic model, although the gap in results is not very large. Finally, regarding the FakeNewsNet dataset, it looks that the deep learning-based model (FakeNewsTracker) does not achieve a good performance comparing to the other proposed baseline by the authors, which is a Logistic Regression (LR) classifier with one-hot vectors of the news articles' text. Furthermore, it seems that a simple word-based model works better than a more sophisticated model that incorporates social media and context information. The FakeFlow model, on the other hand, achieves a better result, outperforming both the FakeNewsTracker and the LR baseline.

\begin{table}
    \footnotesize
    \centering
    \begin{tabular}{|c|c|c|c|c|}
    \hline
    
    \textbf{TruthShades} & Acc. & Prec. & Rec. & F1$_{macro}$ \\ \hline
    \multicolumn{5}{|c|}{\small{Out-of-domain}} \\
    \hline
    {\scriptsize \citet{rashkin2017truth}}  & 0.67 & 0.70 & 0.67 & 0.65 \\ \hline
    FakeFlow                 & 0.68 & 0.69 & 0.68 & \textbf{0.68} \\ \hline
    \hline
    \multicolumn{5}{|c|}{\small{In-domain}} \\
    \hline
    {\scriptsize \citet{rashkin2017truth}}  & 0.91 & 0.91 & 0.91 & 0.91 \\ \hline
    FakeFlow                 & 0.96 & 0.96 & 0.96 & \textbf{0.96} \\ \hline 
    
    \multicolumn{5}{c}{} \\ \hline
    \textbf{PoliticalNews} & Acc. & Prec. & Rec. & F1$_{weighted}$ \\
    \hline
    TopicAgnostic & 0.87 & 0.87 & 0.87 & 0.87 \\ \hline
    FakeFlow      & 0.88 & 0.88 & 0.88 & \textbf{0.88} \\ \hline
    
    \multicolumn{5}{c}{} \\ \hline
    \textbf{FakeNewsNet} & Acc. & Prec. & Rec. & F1$_{weighted}$ \\
    \hline
    FakeNewsTracker     & 0.80 & 0.82 & 0.75 & 0.79 \\ \hline
    One-Hot LR          & 0.82 & 0.90 & 0.72 & 0.80 \\ \hline
    FakeFlow            & 0.86 & 0.86 & 0.86 & \textbf{0.85} \\ \hline
    
    \end{tabular}
    \caption{Results on multiple datasets. We compare the FakeFlow model to SOTA models on each dataset.}
    \label{tab:datasets}
\end{table}

\begin{figure}
\centering
\includegraphics[width=8cm]{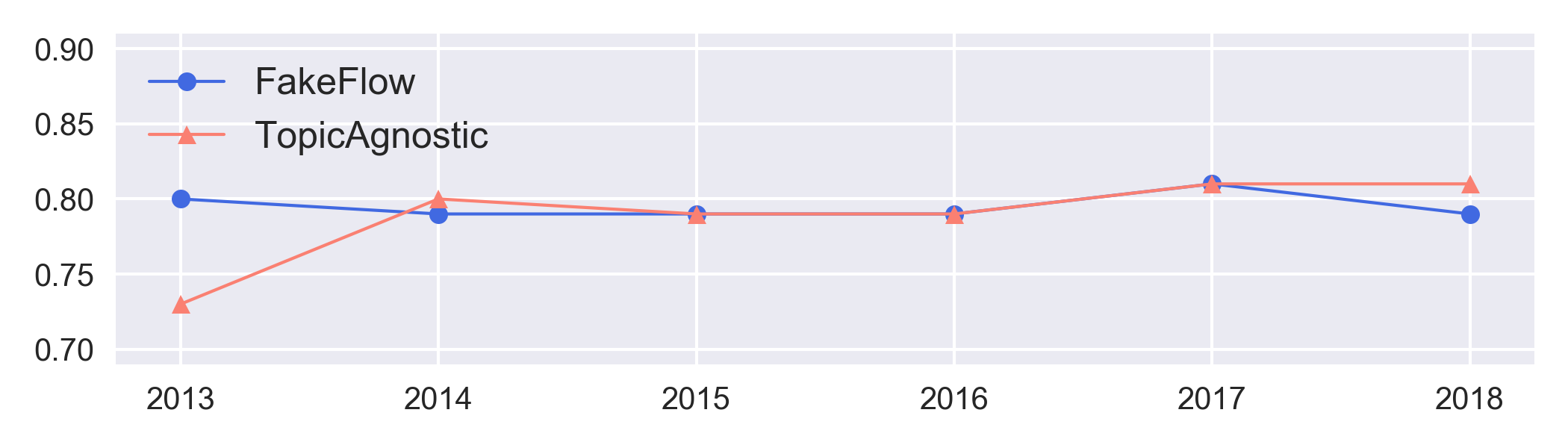}
\caption{Topic aware experiment's results.}
\label{fig:topic_aware}
\end{figure}

\vspace{1em}
\noindent
\textbf{Topic-Aware Model.} Constantly, new events are covered by  news agencies. These events are different from the old ones in terms of discourse and topic. Therefore, a fake news detector trained on news articles from years back is unable to detect recent news. In this experiment, we are evaluating our approach on the PoliticalNews dataset that is constructed from news distributed across different years (2013 to 2018). Following the experimental setup in \cite{castelo2019topic}, we train the FakeFlow model on news from one year and test on the other years, one year at a time for testing. For example, we train the model on news from 2013 and we test on news from 2015. Note that each test set is associated with 5 results, one for each year. 
Figure \ref{fig:topic_aware} shows the average accuracy for each test set. We compare FakeFlow to the TopicAgnostic model that proved to be effective at detecting fake news from different years. It is worth mentioning that the features of the TopicAgnostic model have been extracted from both headlines and text of the news articles. However, the results show that both models have a similar performance, except for the 2013 test set where FakeFlow achieves a higher accuracy with a difference of 7\%. The experiment shows that FakeFlow is capable of detecting fake news from different years, with a flat performance across the years.

\begin{figure*}
\centering
\includegraphics[width=13cm]{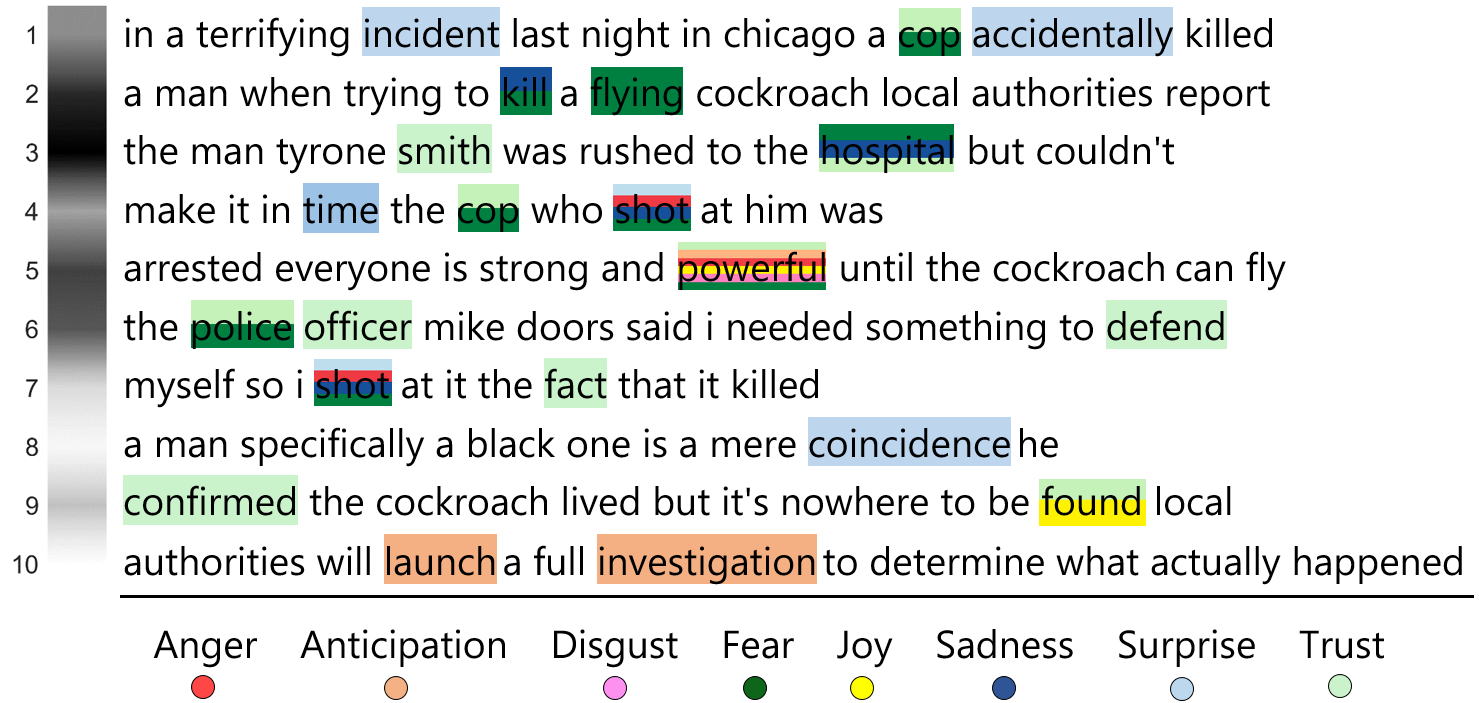}
\caption{Emotional interpretation of a \textit{fake} news article by showing the attention weights (the bar on the left) and highlighting the emotions in the text.}
\label{fig:pipeline}
\end{figure*}

\vspace{1em}
\noindent
\textbf{Attention Weights.} The proposed FakeFlow model shows that taking into account the flow of affective information in fake news is an important perspective for fake news detection. 
We argue that being able to better understand the behaviour of the model can make it more transparent to the end-users. 
Figure \ref{fig:pipeline} illustrates this by showing the attention weights of a fake news article across the 10 segments (left bar).\footnote{We averaged the attention weight matrix along the timesteps (number of segments) representations.} The figure shows that FakeFlow attends more to the beginning of the article. For better understanding, we match the affective information with the attention weights. Regarding the news text in the figure, the \textit{emotions} features\footnote{Words with multiple colors mean that they have been annotated with multiple emotion types in the NRC lexicon.} show a clear example of how fake news articles try to manipulate the reader. It looks as if the existence of \textit{fear, sadness}, and \textit{surprise} \textit{emotions} at the beginning of the article have triggered the attention on this part. Towards the end of the article, on the other hand, we can notice that such negative emotions do not exist, while \textit{emotions} like \textit{joy} and \textit{anticipation} appear. This exemplifies how fake news try to attract the readers' attention in the first part of the text. 
Regarding the \textit{morality} features, we only match the word ``kill'' with the \textit{harm} category. Also, for the \textit{hyperbolic} feature, we match the words ``terrifying'' and ``powerful''. In the same manner, both \textit{morality} and \textit{hyperbolic} features match words that occur at the beginning of the article. Lastly, for both \textit{sentiment} and \textit{imageability} features, we are not able to find a clear interpretation in this example where many words across the segments match.

\begin{figure}[H]
  \begin{minipage}{\textwidth}
    \includegraphics[width=8cm]{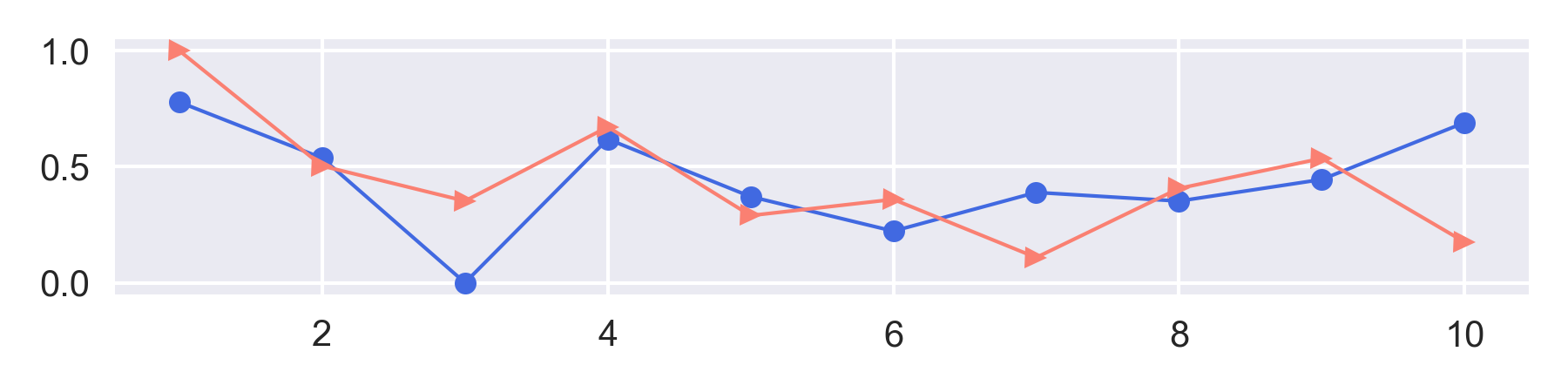}
  \end{minipage}\hfill
    \caption{The flow of the \textit{Fear} emotion in \textbf{fake} (\RIGHTarrow) and \textbf{real} (\textbullet) news articles in the MultiSourceFake dataset. Y-axis presents the average number of \textit{Fear} emotion words in 0-1 scale; the X-axis presents the document text, divided into 10 segments.}
    \label{fig:flow}
\end{figure}

\vspace{1em}
\noindent
\textbf{Real vs.\ Fake Analysis.} In Table \ref{tab:feats_statistics} we present an analysis on both real and fake news articles. The analysis gives an intuition to the reader on the distribution of the used features across the articles' segments. It shows that an emotion like \textit{fear} has on average a higher difference between the first and the last segment in fake news than in real ones (see Figure \ref{fig:flow} for a visualized distribution). Also, a feature like \textit{hyperbolic} has a higher average value and lower standard deviation across all segments for fake news than real news, thus indicating that fake news have a higher amount of hyperbolic words with similarly high values.

\begin{table*}
\footnotesize
    \centering
    \begin{tabular}{c|c|c|c|c|c||c|c|c|c}
    \hline
    & \multirow{2}{*}{\textbf{Features}} & \multicolumn{4}{c|}{\textbf{Real News}} & \multicolumn{4}{|c}{\textbf{Fake News}} \\ \cmidrule{3-10}
    & {} & $\mu$ $first_{seg.}$ & $\mu$ $last_{seg.}$ & $\mu$ $all_{seg.}$ & $\sigma$ $all_{seg.}$ &  $\mu$ $first_{seg.}$ & $\mu$ $last_{seg.}$ & $\mu$ $all_{seg.}$ & $\sigma$ $all_{seg.}$ \\
    \hline
    \parbox[t]{2mm}{\multirow{8}{*}{\rotatebox[origin=c]{90}{Emotions}}} 
    & Anger & 0.175 & 0.167 & 0.170 & 0.003 & 0.183 & 0.170 & 0.171 & 0.008 \\
    & Anticipation & 0.301 & 0.315 & 0.264 & 0.025 & 0.293 & 0.305 & 0.260 & 0.022 \\
    & Disgust & 0.095 & 0.101 & 0.095 & 0.004 & 0.096 & 0.091 & 0.091 & 0.007 \\
    & \bc Fear & \bc 0.254 & \bc 0.250 & \bc 0.238 & \bc 0.010 & \bc 0.265 & \bc 0.226 & \bc 0.238 & \bc 0.011 \\
    & Joy & 0.217 & 0.226 & 0.183 & 0.021 & 0.207 & 0.203 & 0.175 & 0.020 \\
    & Sadness & 0.161 & 0.158 & 0.160 & 0.006 & 0.155 & 0.155 & 0.158 & 0.007 \\
    & Surprise & 0.140 & 0.144 & 0.123 & 0.012 & 0.142 & 0.123 & 0.120 & 0.008 \\
    & Trust & 0.446 & 0.466 & 0.400 & 0.031 & 0.461 & 0.421 & 0.401 & 0.029 \\  \hline
    \parbox[t]{2mm}{\multirow{2}{*}{\rotatebox[origin=c]{90}{Senti.}}} 
    & Positive & 0.599 & 0.623 & 0.558 & 0.030 & 0.608 & 0.591 & 0.554 & 0.032 \\
    & Negative & 0.369 & 0.337 & 0.347 & 0.011 & 0.367 & 0.336 & 0.350 & 0.013 \\ \hline
    \parbox[t]{2mm}{\multirow{10}{*}{\rotatebox[origin=c]{90}{Morality}}}
    & Harm & 0.007 & 0.011 & 0.007 & 0.002 & 0.008 & 0.013 & 0.007 & 0.002 \\
    & Care & 0.026 & 0.023 & 0.019 & 0.004 & 0.021 & 0.022 & 0.019 & 0.003 \\
    & Fairness & 0.003 & 0.013 & 0.007 & 0.002 & 0.005 & 0.020 & 0.009 & 0.004 \\
    & Unfairness & 0.000 & 0.000 & 0.001 & 0.000 & 0.001 & 0.000 & 0.001 & 0.001 \\
    & Loyalty & 0.016 & 0.017 & 0.019 & 0.002 & 0.014 & 0.016 & 0.019 & 0.003 \\
    & Betrayal & 0.004 & 0.003 & 0.005 & 0.001 & 0.002 & 0.003 & 0.004 & 0.001 \\
    & Authority & 0.025 & 0.032 & 0.026 & 0.003 & 0.024 & 0.028 & 0.026 & 0.002 \\
    & Subversion & 0.005 & 0.004 & 0.004 & 0.001 & 0.006 & 0.007 & 0.005 & 0.002 \\
    & Sanctity & 0.005 & 0.005 & 0.004 & 0.001 & 0.005 & 0.006 & 0.005 & 0.002 \\
    & Degradation & 0.003 & 0.004 & 0.003 & 0.001 & 0.006 & 0.004 & 0.003 & 0.001 \\ \hline
    \parbox[t]{2mm}{\multirow{2}{*}{\rotatebox[origin=c]{90}{Img}}} 
    & Imageability & 0.845 & 1.203 & 1.144 & 0.122 & 0.877 & 1.184 & 1.145 & 0.124 \\
    & Abstraction & 0.424 & 0.331 & 0.352 & 0.028 & 0.382 & 0.304 & 0.342 & 0.037 \\
    \hline
    & \bc Hyperbolic & \bc 0.042 & \bc 0.05 & \bc 0.045 & \bc 0.005 & \bc 0.046 & \bc 0.044 & \bc 0.047 & \bc 0.003 
    \end{tabular}
    \caption{ A quantitative analysis of the features existence across articles' segments. We present the average value in the first segment ($\mu$ $first_{seg.}$), the average value in the last segment ($\mu$ $last_{seg.}$), the average value in the all 10 segments ($\mu$ $all_{seg.}$), and the standard deviation ($\sigma$ $all_{seg.}$) of a feature across the 10 segments, both in real and fake news.
    }
    \label{tab:feats_statistics} 
\end{table*}

\section{Conclusion}
In this paper we presented FakeFlow, a model that takes into account the flow of affective information (\textit{emotions, sentiment, hyperbolic words}, etc.) in texts to better detect fake news articles. The model receives as input a text, segmented into smaller units, instead of processing one long fragment. 
This enables it to learn the flow of affective information by modeling the interaction between the topic and affective terms in the news article. 
We evaluated our model on four different datasets and compared it to several strong baselines. The extensive experiments show the effectiveness of FakeFlow over state-of-the-art models. Although FakeFlow was trained using a limited amount of text, the results demonstrated that it achieves results on-par with resource-hungry models (e.g. BERT and Longformer). 
In future work, we plan to extend our dataset and study more fine-grained news types, e.g. propaganda, from an emotional perspective. Moreover, we plan to investigate how we can replace the lexicon-based information with language-independent approaches in an attempt to make our model multilingual.

\section*{Acknowledgment}
The first author would like to thank Ines Rehbein and Ana Uban for their valuable comments and suggestions. The work of the third author was partially funded by the Spanish MICINN under the research project MISMIS-FAKEnHATE on MISinformation and MIScommunication in social media: FAKE news and HATE speech (PGC2018- 096212-B-C31) and by the Generalitat Valenciana under the research project DeepPattern (PROMETEO/2019/121).

\bibliography{eacl2021}
\bibliographystyle{acl_natbib}

\appendix

\section{Appendices}
\label{sec:appendix}
\subsection{Hyper-parameters}
For FakeFlow hyper-parameters, we tune the following parameters with the their correspondent search space:

\begin{itemize}
    \item Dropout: random selection in the range [0.1, 0.6],
    \item Dense layers: [8, 16, 32, 64, 128],
    \item Activation functions: [selu, relu, tanh, elu],
    \item CNN filters' sizes: [(2, 3, 4), (3, 4, 5), (4, 5, 6), (3, 5), (2, 4), (4,), (5,), (3, 5, 7), (3, 6)],
    \item Numbers of CNN filters: [4, 8, 16, 32, 64, 128],
    \item Pooling size: [2, 3],
    \item GRU units: [8, 16, 32, 64, 128],
    \item Optimization function: [adam, adadelta, rmsprop, sgd],
    \item For the early stopping, we set the `patience` parameter to 4 and we set the epochs number to 50.
\end{itemize}

For the parameters selection, we use hyperopt\footnote{https://github.com/hyperopt/hyperopt} library that receives the above search space to randomly select different $N$ combination of parameters (trials). We use a small value of $N$ in all of our experiments to avoid overdrawn finetuning; we set $N$ to 35.

\subsection{Topic Aware experiments}
In Figure 4, we present the average accuracy of our model when we train on different years and test a specific one. In the following we show the results before we averaged them.
\begin{table}[H]
    \small
    \centering
    \begin{tabular}{|c|c|c|c|c|c|c|}
    \hline
    \backslashbox{Train}{Test} & 2013 & 2014 & 2015 & 2016 & 2017 & 2018 \\
    \hline
    2013        & 0.00 & 0.82 & 0.74 & 0.76 & 0.78 & 0.74 \\ \hline
    2014        & 0.84 & 0.00 & 0.79 & 0.76 & 0.81 & 0.74 \\ \hline
    2015        & 0.79 & 0.81 & 0.00 & 0.82 & 0.80 & 0.82 \\ \hline
    2016        & 0.80 & 0.76 & 0.87 & 0.00 & 0.85 & 0.79 \\ \hline
    2017        & 0.79 & 0.82 & 0.76 & 0.80 & 0.00 & 0.85 \\ \hline
    2018        & 0.79 & 0.75 & 0.81 & 0.83 & 0.83 & 0.00 \\ \hline \hline
    Average     & 0.80 & 0.79 & 0.79 & 0.79 & 0.81 & 0.79 \\ \hline
    \end{tabular}
    \caption{FakeFlow results for each train-test run for the Topic-Aware experiment.}
\end{table}

\end{document}